\documentclass{article}

\usepackage{amssymb}
\usepackage{bm}
\usepackage{graphicx}
\PassOptionsToPackage{hyphens}{url}\usepackage{hyperref}

\newcommand{\x}{\mathbf{x}}
\renewcommand{\u}{\mathbf{u}}
\newcommand{\y}{\mathbf{y}}
\newcommand{\W}{\mathbf{W}}

\newcommand{\Wh}{\hat{\W}}
\newcommand{\Wout}{\W_{out}}
\newcommand{\R}{\mathbb{R}}

\begin{document}

\title{Deep Echo State Network (DeepESN):\\ A Brief Survey}

\author{Claudio Gallicchio and Alessio Micheli\\
\small{Department of Computer Science, University of Pisa}\\
\small{gallicch@di.unipi.it, micheli@di.unipi.it}
}
\date{}
\maketitle

\begin{abstract}
The study of deep recurrent neural networks (RNNs) and, in particular, of deep Reservoir Computing (RC) is gaining an increasing research attention in the neural networks community.
The recently introduced Deep Echo State Network (DeepESN) model opened the way to an extremely efficient approach for designing deep neural networks for temporal data. At the same time, the study of DeepESNs allowed to shed light on the intrinsic properties of state dynamics developed by hierarchical compositions of recurrent layers, i.e. on the bias of depth in 
RNNs architectural design.
In this paper, we summarize the advancements in the development, analysis and applications of DeepESNs.

\vspace{0.2cm}

\end{abstract}

\section{Introduction}
In the last decade, the Reservoir Computing (RC) paradigm \cite{Verstraeten2007,Lukosevicius2009}
has attested as a state-of-the-art approach for
the design of efficiently trained Recurrent Neural Networks (RNNs).
Though different instances of the RC methodology exist in literature (see e.g. \cite{Maass2002real,Tino2001predicting}),
the Echo State Network (ESN) \cite{Jaeger2004,Jaeger2001} certainly represents the most widely known model, with a strong theoretical ground (e.g. \cite{Tino2007,Gallicchio2011NN,Yildiz2012re,Manjunath2013echo,Massar2013mean,Tino2017ESANN})
and a plethora of successful applications reported in literature (see e.g. \cite{lukovsevivcius2012reservoir,schrauwen2007overview} and references therein,
as well as more recent works e.g. in \cite{Gallicchio2016Human,Gallicchio2015Electricity,Gallicchio2014NCA}).
Essentially, ESNs are recurrent randomized neural networks \cite{Gallicchio2018Randomized,Gallicchio2017Randomized} in which the state dynamics are implemented by an untrained recurrent hidden layer, whose activation is used to feed a static output module that is the only trained part of the network.
In this paper we deal with the extension of the ESN approach to the deep learning 
framework.
This line of research can be interestingly framed within the context of deep randomized neural networks \cite{gallicchio2020deep}, in which the analysis is focused on the behavior of deep neural architectures where most of the connections are untrained. 

The study of deep neural network architectures for temporal data processing is
an attractive area of research in the neural networks community \cite{Angelov2016, Goodfellow-et-al-2016-Book}. 
Investigations in the field of hierarchically organized Recurrent Neural Networks (RNNs) showed that deep RNNs are able to develop in their internal states a multiple time-scales representation of the temporal information, a much desired feature e.g. when approaching complex tasks in the area of speech or text processing \cite{Graves2013speech,Hermans2013}.

Recently, the introduction of the Deep Echo State Network (DeepESN) model \cite{Gallicchio2017DeepESN,Gallicchio2016DeepESANN} allowed to study the properties of layered RNN architectures separately from the learning aspects. Remarkably, such studies pointed out that the structured state space organization with multiple time-scales dynamics in deep RNNs is intrinsic to the nature of compositionality of recurrent neural modules. 
The interest in the study of the DeepESN model is hence twofold. On the one hand,  
it allows to shed light on the intrinsic properties of state dynamics of layered RNN architectures
\cite{Gallicchio2018layering}. On the other hand it enables the design of extremely efficiently trained deep neural networks for temporal data.

Previous to the explicit introduction of the DeepESN model in \cite{Gallicchio2017DeepESN}, 
works on hierarchical RC models targeted ad-hoc constructed architectures, where different modules were trained 
for discovery of temporal features at different scales on synthetic data \cite{Jaeger2007discovering}.
Ad-hoc constructed modular networks made up of multiple ESN modules have also been investigated in the speech processing area  
\cite{Triefenbach2013acoustic,Triefenbach2010phoneme}.
More recently, the advantages of multi-layered RC networks have been experimentally studied on time-series
benchmarks in the RC area \cite{Malik2017multilayered}.
Differently from the above mentioned works, the studies on DeepESN considered in the following aim to address some fundamental questions pertaining to the true nature of layering as a factor of architectural RNN design \cite{Gallicchio2018layering}.
Such basic questions can be essentially summarized as follows:
(i)  Why stacking layers of recurrent units?
(ii) What is the inherent architectural effect of layering in RNNs (independently from learning)?
(iii) Can we extend the advantages of depth in RNN design using efficiently trained RC approaches?
(iv) Can we exploit the insights from such analysis to address the automatic design of deep recurrent models 
(including fundamental parameters such as the architectural form, the number of layers, the number of units in each layer, etc.)?

This paper is intended both to draw a line of recent developments in response to the above mentioned key research questions
and  to provide an up-to-date overview  on the progress and on the perspectives in the studies of DeepESNs,
which are presented in Section \ref{sec.Advances}. Before that, in Section~\ref{sec.deepESN} we recall the major characteristics of the DeepESN model.

\section{The Deep Echo State Network Model}
\label{sec.deepESN}

\begin{figure}[tbh]
	\centering
		\includegraphics[width=1\textwidth]{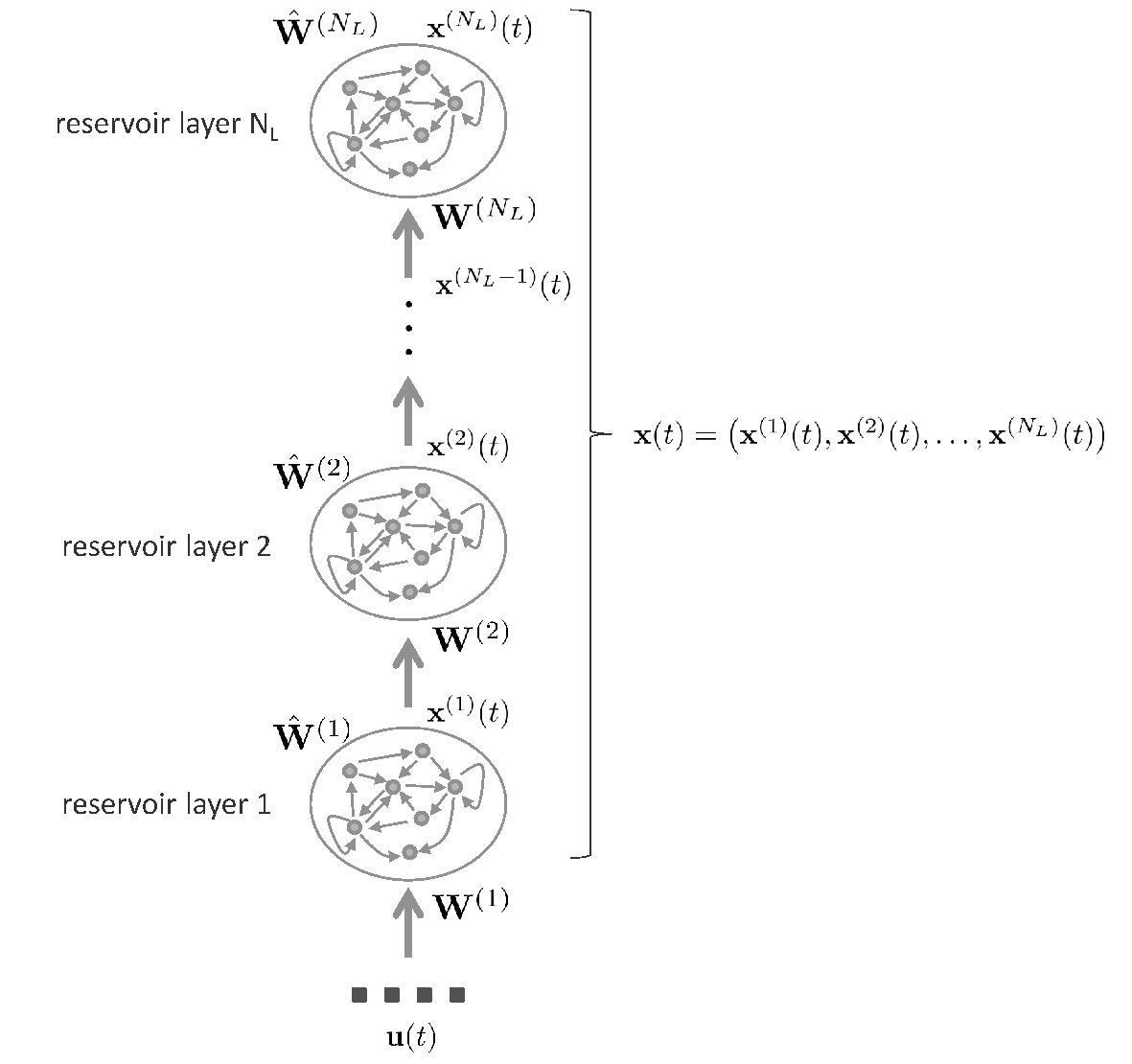}
	\caption{Reservoir architecture of a Deep Echo State Network.}
	\label{fig.architecture}
\end{figure}

As for the standard shallow ESN model, a DeepESN is composed by a dynamical \emph{reservoir} component, which embeds the input history into a rich state representation,
and by a feed-forward \emph{readout} part, wich exploits the state econding provided by the reservoir to compute the output.
Crucially, the reservoir of a DeepESN is organized into a \emph{hierarchy of stacked recurrent layers}, where the output of each layer acts as input for the next one,
as illustrated in Figure \ref{fig.architecture}. 

In this case, at each time step $t$, the state computation proceeds by following the pipeline of recurrent layers, from the first one, which is directly fed by the external input, up to the highest one in the reservoir architecture. In our notation we use $N_U$ to denote the external input dimension, $N_L$ to indicate the number of reservoir layers, and we assume, for the sake of simplicity, that each reservoir layer has $N_R$ recurrent units.
Moreover, we use $\u(t) \in \R^{N_U}$ to denote the external input at time step $t$, while $\x^{(i)}(t) \in 
\R^{N_R}$ is the state of the reservoir layer $i$ at time step $t$. In general, we use the superscript $(i)$ 
to indicate that an item is related to the $i$-th reservoir in the stack.
At each time step $t$, the composition of the states in all the reservoir layers, i.e.
$\x(t) = \big( \x^{(1)}(t), \ldots, \x^{(N_L)}(t) \big) \in \R^{N_R\,N_L}$, gives the global state of the network.

The computation carried out by the stacked reservoir of a DeepESN can be understood under a dynamical system viewpoint as an input-driven discrete-time non-linear dynamical system, where the evolution of the global state $\x(t)$ is governed by a state transition function $F = \big(F^{(1)}, \ldots, F^{(N_L)}\big)$,
with each $F^{(i)}$ ruling the state dynamics at layer $i$.
Assuming leaky integrator reservoir units \cite{Jaeger2007} in each layer and omitting the bias terms for the ease of notation, the reservoir dynamics of a DeepESN are mathematically described as follows.
For the first layer we have that:
\begin{equation}
\label{eq.layer1}
\begin{array}{ll}
\x^{(1)}(t) & =  F(\u(t),\x^{(1)}(t-1))\\
& = (1-a^{(1)}) \x^{(1)}(t-1) + a^{(1)} \mathbf{f}(\W^{(1)} \u(t) + \Wh^{(1)} \x^{(1)}(t-1)),
\end{array}
\end{equation}
while for successive layers $i>1$ the state update is given by:
\begin{equation}
\label{eq.layeri}
\begin{array}{ll}
\x^{(i)}(t) & =  F(\x^{(i-1)}(t),\x^{(i)}(t-1))\\
& = (1-a^{(i)}) \x^{(i)}(t-1) + a^{(i)} \mathbf{f}(\W^{(i)} \x^{(i-1)}(t) + \Wh^{(i)} \x^{(i)}(t-1)).
\end{array}
\end{equation}
In the above equations~\ref{eq.layer1} and \ref{eq.layeri}, 
$\W^{(1)} \in \R^{N_R \times N_U}$ is the input weight matrix,
$\W^{(i)} \in \R^{N_R \times N_R}$ for $i>1$ is the weight matrix for inter-layer connections from layer $(i-1)$
to layer $i$, $\Wh^{(i)} \in \R^{N_R \times N_R}$ is the recurrent weight matrix for layer $i$,
$a^{(i)} \in [0,1]$ is the leaking rate for layer $i$ and $\mathbf{f}$ denotes the element-wise applied activation
function for the recurrent reservoir units (typically, the $tanh$ non-linearity is used).

\begin{figure}[tb]
	\centering
		\includegraphics[width=0.39\textwidth]{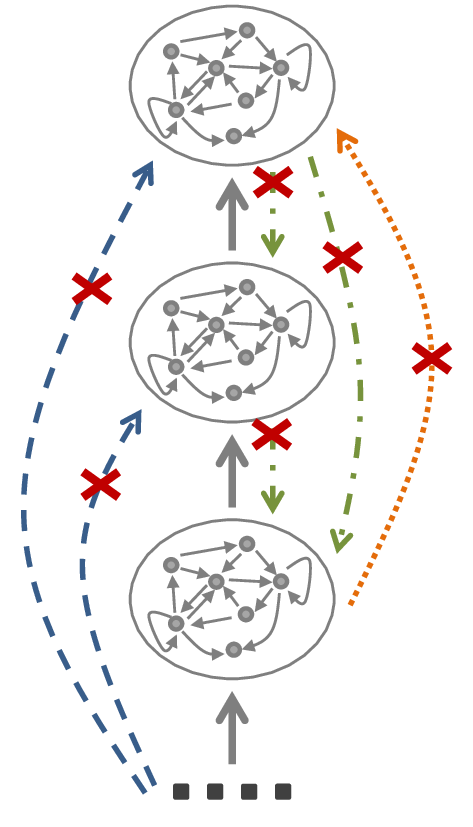}
	\caption{The layered reservoir architecture of DeepESN as a constrained version of a shallow reservoir.
	Compared to the shallow case with the same total number of recurrent units, 
	in a stacked DeepESN architecture the following connections are removed:
	from the input to reservoir levels at height $>$ 1 (blue dashed arrows),
	from higher to lower reservoir levels (green dash dotted arrows),
	from each reservoir at level $i$ to all reservoirs at levels $> i+1$ (orange dotted arrows).}
	\label{fig.constraints}
\end{figure}

Interestingly, as graphically illustrated in Figure~\ref{fig.constraints}, we can observe that the reservoir architecture of a DeepESN can be characterized, with respect to the shallow counterpart, by interpreting it as a constrained version of standard shallow ESN/RNN with the same total number of recurrent units.
In particular, the following constraints are applied in order to obtain a layered architecture:
\begin{itemize}
\item all the connections from the input layer to reservoir layers at a level higher than 1 are removed (influencing the way in which the external input information is seen by recurrent units progressively more distant from the input layer);
\item all the connections from higher layers to lower ones are removed (which affects the flow of information and the dynamics of sub-parts of the network's state);
\item all the connections from each layer to higher layers different from the immediately successive one in the pipeline are removed (which affects the flow of information and the dynamics of sub-parts of the network's state).
\end{itemize}
The above mentioned constraints, that graphically correspond to layering, have been explicitly and extensively discussed in our previous work in \cite{Gallicchio2017DeepESN}. 
Under this point of view, the DeepESN architecture can be seen as a simplification of the corresponding single-layer ESN, 
leading to a reduction in the absolute number of recurrent weights which, assuming full-connected reservoirs at each layer,
is quadratic in both the number of recurrent units per layer and total number of layers \cite{Gallicchio2018Lyapunov}.
As detailed in the above points, however, note that this peculiar architectural organization influences the way in which the temporal information is processed by the different sub-parts of the hierarchical reservoir, composed by recurrent units that are progressively more distant from the external input.

Furthermore, differently from the case of a standard ESN/RNN,
the state information transmission between consecutive layers in a DeepESN presents no temporal delays. In this respect, we can make the following considerations:
\begin{itemize}
\item the aspect of sequentiality between layers operation is already present and discussed in previous works in literature on deep RNN (see e.g. \cite{Hermans2013,Graves2013speech,El1996hierarchical,Schmidhuber1992learning}), which actually stimulated the investigation on the intrinsic role of layering in such hierarchically organized recurrent network architectures;
\item this choice allows the model to process the temporal information at each time step in a ``deep'' temporal fashion, i.e. through a hierarchical composition of multiple levels of recurrent units;
\item in particular, notice that the use of (hyperbolic tangent) non-linearities applied individually to each layer during the state computation does not allow to describe the DeepESN dynamics by means of an equivalent shallow system.
\end{itemize}

Based on the above observations, a major research question naturally arises and drives the motivation to the studies reported in Section~\ref{sec.Advances}, i.e. how and to what extent the described constraints that rule the layered construction 
and the hierarchical representation in deep recurrent models have an influence on their dynamics.

As in the standard RC framework, the reservoir parameters, i.e. the weights in matrices $\W^{(i)}$ and $\Wh^{(i)}$, are left untrained after initialization under stability constraints, which are given through the analysis of the Echo State Property for deep reservoirs
provided in \cite{Gallicchio2017echo}.

\begin{figure}[thb]
	\centering
		\includegraphics[width=0.45\textwidth]{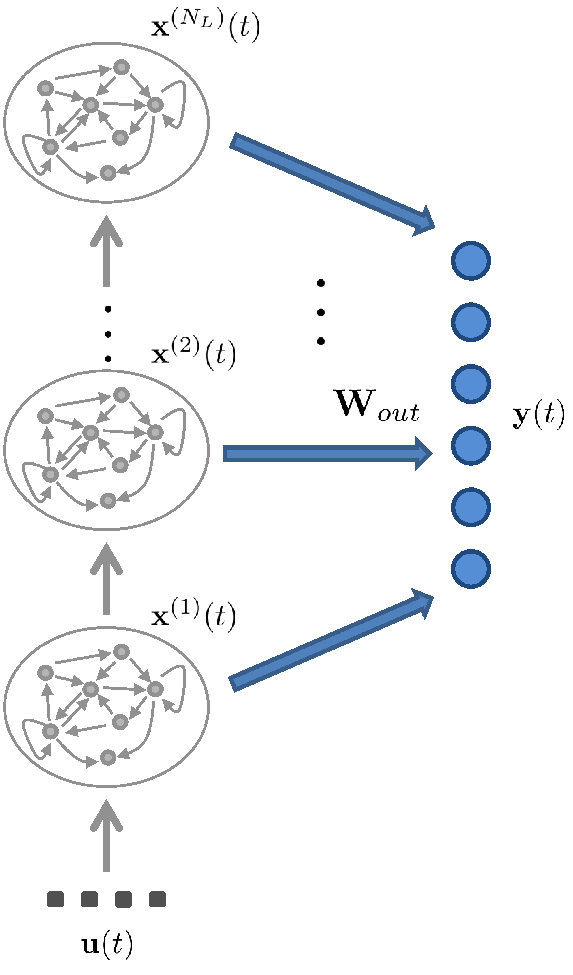}
	\caption{Readout organization for DeepESN 
	in which at each time step the reservoir states of all  layers are used as input for the output layer.}
	\label{fig.readout}
\end{figure}

As regards the output computation, although different choices are possible for the pattern of connectivity between the reservoir layers and the output module (see e.g. \cite{Hermans2013,Pascanu2014}), a typical setting consists in 
feeding at each time step $t$ the state of all reservoir layers (i.e. the global state of the DeepESN) to the output layer,
as illustrated in Figure~\ref{fig.readout}.
Note that this choice enables the readout component to give different weights to the dynamics developed at different layers, thereby allowing to exploit the potential variety of state representations in the stacked reservoir.
Denoting by $N_Y$ the size of the output space, in the typical case of linear readout,
the output at time step $t$ is computed as:
\begin{equation}
\y(t) = \Wout \x(t) = \Wout \big( \x^{(1)}(t),\ldots, \x^{(N_L)}(t) \big),
\end{equation}
where  $\Wout \in \R^{N_L \times N_R\,N_L}$ 
is the readout weight matrix that is adapted on a training set, typically in closed form through
direct methods such as pseudo-inversion or ridge-regression.

\section{Advances}
\label{sec.Advances}
Here we briefly survey the recent advances in the study of the DeepESN model. 
The works described in the following, by addressing the key questions summarized in the Introduction,
provide a general support to the significance of the DeepESN, also critically discussing advantages and drawbacks 
of its construction.

\begin{itemize}
\item The DeepESN model has been introduced in \cite{Gallicchio2017DeepESN}, which extends the preliminary work in \cite{Gallicchio2016DeepESANN}.
The analysis provided in these papers revealed, through empirical investigations, 
the hierarchical structure of temporal data representations developed by the layered reservoir architecture of a DeepESN.
Specifically, the stacked composition of recurrent reservoir layers was shown to enable a \emph{multiple time-scales representation} of the temporal information, naturally ordered along the network's hierarchy.
Besides, in \cite{Gallicchio2017DeepESN} layering proved effective also as a way to enhance the effect of known RC factors of network design, including unsupervised reservoir adaptation by means of Intrinsic Plasticity \cite{Schrauwen2008}.
The resulting effects have been analyzed also in terms of state entropy and memory.

\item The hierarchically structured state representation in DeepESNs has been investigated by means of \emph{frequency analysis} in \cite{Gallicchio2017hierarchicalWIRN}, which specifically considered the case of recurrent units with 
\emph{linear} activation functions.
Results pointed out the intrinsic multiple frequency representation in DeepESN states, 
where, even in the simplified linear setting, progressively higher layers focus on progressively lower frequencies.
In \cite{Gallicchio2017hierarchicalWIRN} the potentiality of the deep RC approach has also been exploited 
in predictive experiments, showing that DeepESNs outperform state-of-the-art results on the 
class of Multiple Superimposed Oscillator (MSO) tasks by several orders of magnitude.

\item The fundamental RC conditions related to the \emph{Echo State Property} (ESP) have been generalized to the case of deep RC networks in \cite{Gallicchio2017echo}. Specifically, through the study of stability and contractivity  of nested dynamical systems, the theoretical analysis in \cite{Gallicchio2017echo} gives a sufficient condition and a necessary condition for the Echo State Property to hold in case of deep RNN architectures.
Remarkably, the work in \cite{Gallicchio2017echo} provides a relevant conceptual and practical tool for the definition, validity and usage of DeepESN in an ``autonomous'' way with respect to the standard ESN model.

\item The study of DeepESN dynamics under a dynamical system perspective has been pursued in \cite{Gallicchio2018Lyapunov,Gallicchio2017LyapunovESANN}, which provide a theoretical and practical framework for the study of stability of layered recurrent dynamics in terms of \emph{local Lyapunov exponents}.
This study also provided interesting insights 
in terms of the quality of the developed system dynamics, showing that (under simple initialization settings) layering has the effect of naturally pushing the global dynamical regime of the recurrent network closer to the stable-unstable transition condition known as the edge of chaos \cite{bertschinger2004real,legenstein2007edge,boedecker2012information}.

\item The study of the frequency spectrum of deep reservoirs enabled to address one of the fundamental open issues in deep learning, namely \emph{how to choose the number of layers in a deep RNN architecture}.
Starting from the analysis of the intrinsic differentiation of the filtering effects of successive levels in a stacked RNN architecture, the work in \cite{Gallicchio2018design} proposed an automatic method for the \emph{design} of DeepESNs. Noticeably, the proposed approach allows to tailor the DeepESN architecture to the characteristics of the input signals, 
consistently relieving the cost of the model selection process, and leading to new state-of-the-art results in speech and music processing tasks.

\item A first extension of the deep RC framework for \emph{learning in structured domains} has been presented in \cite{Gallicchio2019deep, Gallicchio2018DeepTESN},
which introduced the Deep Tree Echo State Network (DeepTESN) model. 
The new model points out that it is possible to combine the concepts of deep learning, learning for trees and RC training efficiency, taking advantages from the layered architectural organization and from the compositionality of the structured representations both in terms of efficiency and in terms of effectiveness.
On the application side, the experimental results reported in \cite{Gallicchio2019deep, Gallicchio2018DeepTESN} concretely showed that the deep RC approach for trees can be extremely advantageous, beating previous state-of-the-art results in challenging tasks from domains of document processing and computational biology.
As regards the mathematical description of the model, the reservoir operation is extended to implement a (non-linear) state transition system over discrete tree structures, whose asymptotic stability analysis enables the definition of a generalization of the ESP for the case of tree structured data \cite{Gallicchio2019deep}.
Overall, DeepTESN provides a first instance of an extremely efficient approach for the design of deep neural networks for learning in cases where the input is given by (hierarchically-)structured data.
Moreover, from a theoretical viewpoint, the work in \cite{Gallicchio2019deep} 
also gives an in-depth analysis of asymptotic stability of untrained (and non-linear) state transition systems operating on discrete trees. 
In this context, the analysis in \cite{Gallicchio2019deep}
results into a generalization of the ESP of conventional reservoirs, described under the name of Tree Echo State Property.

The Deep RC approach has been proved very advantageous also in the case of learning with graph data, enabling the development of Fast and Deep Graph Neural Networks (FDGNNs) in \cite{gallicchio2020fast}. 
The concept of reservoirs operating on discrete graph structures has been first introduced in \cite{gallicchio2010graph}, and revolves around the computation of a state embedding for each vertex in an input graph. In particular, the state for a vertex $v$ is computed as a function of the input information attached to the vertex $v$ itself (i.e., a vector of features that takes the role of external input in the system), and of the state computed for the neighbors of $v$ (a concept that takes the role of ``previous time-step'' in the case of conventional RC systems for time-series). The stability of the resulting dynamics can be studied by generalizing the mathematical means 
considered for conventional reservoirs,
leading to the definition of Graph Embedding Stability (GES), a stability property for neural embedding systems on graphs introduced in \cite{gallicchio2020fast}, to which the interested reader is referred for further information. Besides the introduction of GES, the work in \cite{gallicchio2020fast} shows how to design a deep RC system for graphs, where each layer builds its embedding on the basis of the state information produced in the previous layer. The FDGNN approach was shown to reach (and even outperform) state-of-the-art accuracy on known benchmarks for graph classification, comparing well with many literature approaches, especially based on convolutional neural networks and kernel for graphs. Inheriting the easy of training algorithms from the RC paradigm, the approach is also extremely faster than literature models, enabling a sensible speed-up in the required training times (up to $\approx$ 3 orders of magnitude in the experiments reported in \cite{gallicchio2020fast}).

\item 
For what regards the experimental analysis in \emph{applications}, DeepESNs were shown to bring several advantages in both cases of synthetic and real-world tasks. Specifically, DeepESNs outperformed shallow reservoir architectures (under fair conditions on the number of total recurrent units and, as such, on the number of trainable readout parameters) on the Mackey-Glass next-step prediction task \cite{Gallicchio2018layering}, on the short-term Memory Capacity task \cite{Gallicchio2017DeepESN,Gallicchio2018MC}, on MSO tasks \cite{Gallicchio2017hierarchicalWIRN}, as well as on a Frequency Based Classification task \cite{Gallicchio2018design}, purposely designed to assess multiple-frequency representation abilities.
As pertains to \emph{real-world problems}, the DeepESN approach recently proved effective in a variety of domains, including Ambient Assisted Living (AAL) \cite{Gallicchio2017AI*AAL}, medical diagnosis \cite{Gallicchio2018Parkinson}, speech and polyphonic music processing \cite{Gallicchio2018design,Gallicchio2019comparison},
metereological forecasting
\cite{alizamir2020deep,kim2020time},
solar irradiance prediction
\cite{li2020multi}, energy consumption and wind power generation prediction \cite{hu2020forecasting},
short-term traffic forecasting  \cite{del2020deep}, destination prediction 
\cite{song2020destination}
car parking and bike-sharing 
in urban computing \cite{kim2020time},
financial market predictions \cite{kim2020time},
and
industrial applications (for blast furnace off-gas) \cite{dettori2020deep,colla2019reservoir}.

\item
\emph{Software} implementations of the DeepESN model have been recently made publicly available in the following forms:
\begin{itemize}
\item DeepRC TensorFlow Library (DeepRC-TF)\\
\small
\url{https://github.com/gallicch/DeepRC-TF}.
\normalsize
\item DeepESN Python Library (DeepESNpy)\\
\small 
\url{https://github.com/lucapedrelli/DeepESN.}
\normalsize
\item Deep Echo State Network (DeepESN) MATLAB Toolbox\\
\small
\url{https://it.mathworks.com/matlabcentral/fileexchange/69402-deepesn.}
\normalsize
\item Deep Echo State Network (DeepESN) Octave library\\
\small
\url{https://github.com/gallicch/DeepESN_octave.}
\normalsize
\end{itemize}
Please note that references \cite{Gallicchio2017DeepESN,Gallicchio2018design} represent citation requests for the use of the above mentioned libraries.
\end{itemize}

\section{Conclusions}

In this survey we have provided a brief overview of the extension of the RC approach towards the deep learning framework, describing the salient features of the DeepESN model.
Noticeably, DeepESNs enable the analysis of the intrinsic properties of state dynamics in deep RNN architectures,
i.e. the study of the bias due to layering in the design of RNNs.
At the same time, DeepESNs allow to transfer the striking advantages of the ESN methodology to the case of deep recurrent  architectures, leading to an efficient approach for designing deep neural networks for temporal data.

The analysis of the distinctive characteristics and dynamical properties of the DeepESN model has been carried out  
first empirically, in terms of entropy of state dynamics and system memory. Then, it has been conducted through more abstract theoretical investigations that allowed the derivation of the fundamental conditions for the ESP of deep networks, 
as well as the characterization of the developed dynamical regimes in terms of local Lyapunov exponents.
Besides, studies on the frequency analysis of DeepESN dynamics allowed us to develop 
an algorithm for the automatic setup of (the number of layers of) a DeepESN.
Current developments already include  model variants and 
applications to both synthetic and real-world tasks. 
Finally, a pioneering extension of the deep RC approach to learning in structured domains has been introduced
with DeepTESN (for trees) and FDGNN (for graphs).

Overall, the final aim of this paper is to summarize the successive advances in the development, analysis and applications of the DeepESN model, providing a document that is intended to contain a constantly updated view over this research topic.

\bibliographystyle{elsarticle-num} 
\bibliography{references}

\end{document}